\pdfoutput=1

\documentclass[11pt]{article}

\usepackage[]{EACL2023}

\usepackage{times}
\usepackage{latexsym}
\usepackage{alltt}
\usepackage{caption}
\usepackage{graphicx}
\usepackage{amssymb}
\usepackage{amsmath}
\usepackage{amsfonts}
\usepackage{pifont}

\usepackage[T1]{fontenc}

\usepackage[utf8]{inputenc}

\usepackage{microtype}
\usepackage{multirow}
\usepackage{comment}

\usepackage{inconsolata}

\title{Investigating the Effect of Relative Positional Embeddings on AMR-to-Text Generation with Structural Adapters}

\author{Sebastien Montella\Thanks{Work done while at Orange Innovation (Lannion, France). Now at Huawei Edinburgh Research Centre (United Kingdom).} \\
 Orange Innovation / Lannion, France \\
  Aix-Marseille Univ. CNRS, LIS / Marseille, France \\
  \texttt{sebastien.montella@huawei.com} \\
  \AND
  Alexis Nasr \\ 
  AMU/CNRS/LIS, Marseille, France \\
  \texttt{alexis.nasr@lis-lab.fr} \\
  \And Johannes Heinecke \\ 
  Orange Innovation / Lannion, France \\
  \texttt{johannes.heinecke@orange.com} \\
  \AND Frederic Bechet \\
  AMU/CNRS/LIS, Marseille, France \\
  \texttt{frederic.bechet@lis-lab.fr} \\
  \And
  Lina M. Rojas-Barahona \\
  Orange Innovation / Lannion, France \\
  \texttt{linamaria.rojasbarahona} \\
  \texttt{@orange.com}
 }

\begin{document}
\maketitle
\begin{abstract}

Text generation from Abstract Meaning Representation (AMR) has substantially benefited from the popularized Pretrained Language Models (PLMs). 
Myriad approaches have linearized the input graph as a sequence of tokens to fit the PLM tokenization requirements. Nevertheless, this transformation jeopardizes the structural integrity of the graph and is therefore detrimental to its resulting representation. To overcome this issue, \citet{ribeiro-etal-2021-structural} have recently proposed StructAdapt, a structure-aware adapter which injects the input graph connectivity within PLMs using Graph Neural Networks (GNNs). In this paper, we investigate the influence of Relative Position Embeddings (RPE) on AMR-to-Text, and, in parallel, we examine the robustness of StructAdapt. Through ablation studies, graph attack and link prediction, we reveal that RPE might be partially encoding input graphs. We suggest further research regarding the role of RPE will provide valuable insights for Graph-to-Text generation.

\end{abstract}

\section{Introduction}
Earliest works on AMR-to-Text generation  were mostly based on statistical methods. A common practice was to convert AMR-to-Text task into an already studied problems such as Tree-to-Text \cite{flanigan-etal-2016-generation, lampouras-vlachos-2017-sheffield}, aligned text-to-text \cite{pourdamghani-etal-2016-generating}, Travel Sales Problems \cite{song-etal-2016-amr} or Grammatical Framework \cite{grammatical-framework-book-ranta}. Recently, most methods are neural-centered with an encoder-decoder architecture \cite{sutskever-etal-2014-seq2seq} as a backbone \cite{konstas-etal-2017-neural, takase-etal-2016-neural, cao-clark-2019-factorising}. Unfortunately, this architecture coerces the AMR to be linearized as a sequence of tokens. This ends up in structural information loss. To tackle this issue, several strategies have attempted to integrate structure using message propagation \cite{song-etal-2018-graph, guo-etal-2019-densely,  damonte-cohen-2019-structural, ribeiro-etal-2019-enhancing, zhang-etal-2020-lightweight, zhao-etal-2020-line}. A limitation of those is the absence of pretraining, as demonstrated by \citet{ribeiro-etal-2021-investigating}. To this end, \citet{ribeiro-etal-2021-structural} introduced StructAdapt for lightweight AMR-to-Text with structural adapters. As linearization and tokenization of the input graph are mandatory steps for PLMs, StructAdapt first defines a new graph where nodes are the resulting subwords from the tokenization. As a result, adapter can henceforth include GNN layers operating on the subsequent graph while leveraging pretrained representations.

However, although studies have been made to probe position embeddings \cite{wang-chen-2020-position, wang-et-al-2021-APE-BERT,dufter-etal-2022-position}, their role on graph encoding has remained unanswered. 
In this paper, we are particularly interested in measuring the saliency of RPE with StructAdapt for AMR-to-Text generation. Our novelty is not in proposing a new method to encode graphs such as \cite{schmitt-etal-2021-modeling} but rather in revealing the interesting behaviours of RPE along with StructAdapt.

\section{\textsc{StructAdapt}: A Structural Adapter}
\begin{figure}[t]
    \centering
    \includegraphics[width=\linewidth]{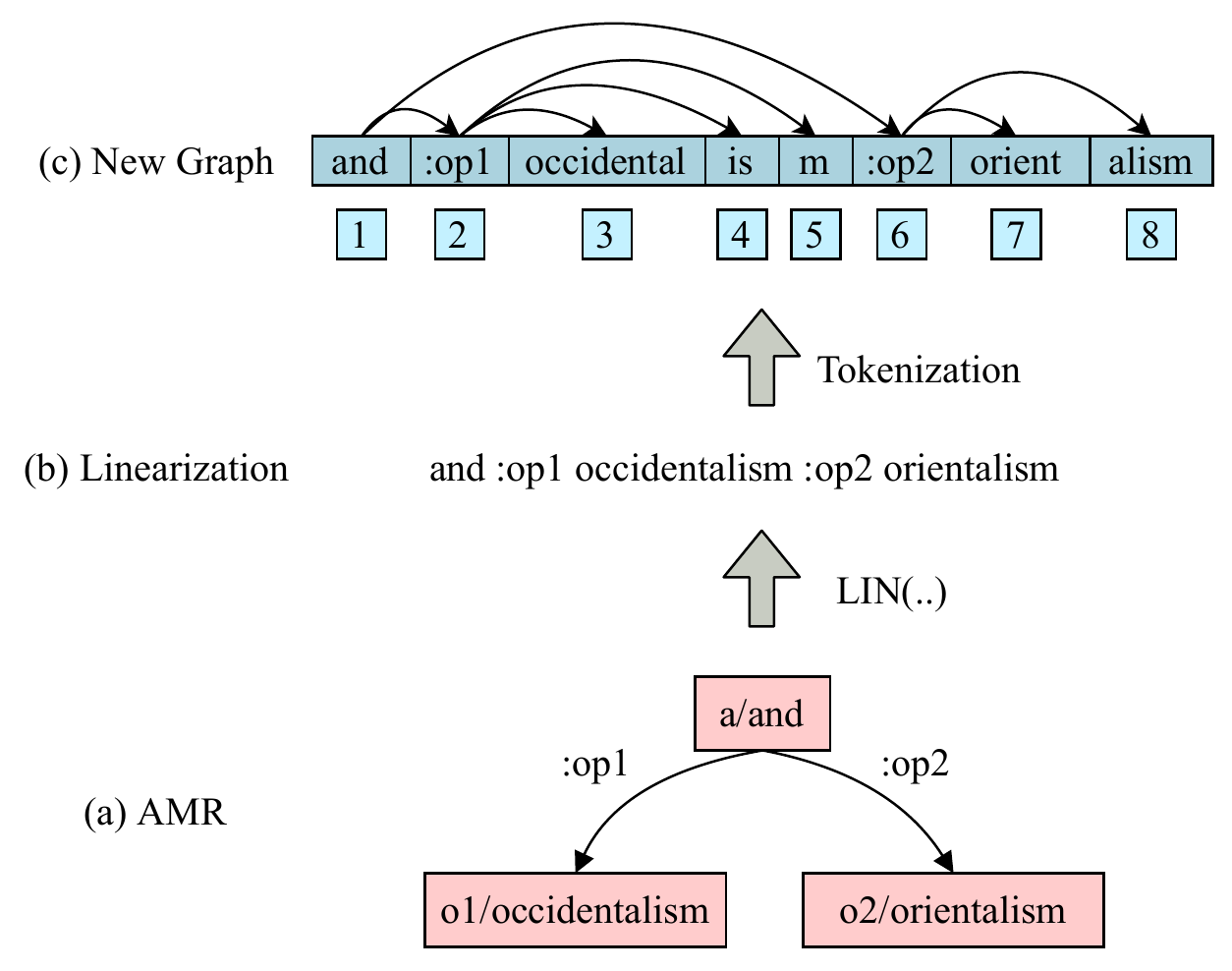}
    \caption{Examples of AMR tokenization for the sentence \textit{``Occidentalism and Orientialism.''}. The resulting input graph in (c) contains 8 nodes.}
    \label{fig:tokenized_amr}
\end{figure}

\begin{figure}[h]
    \centering
    \includegraphics[width=\linewidth]{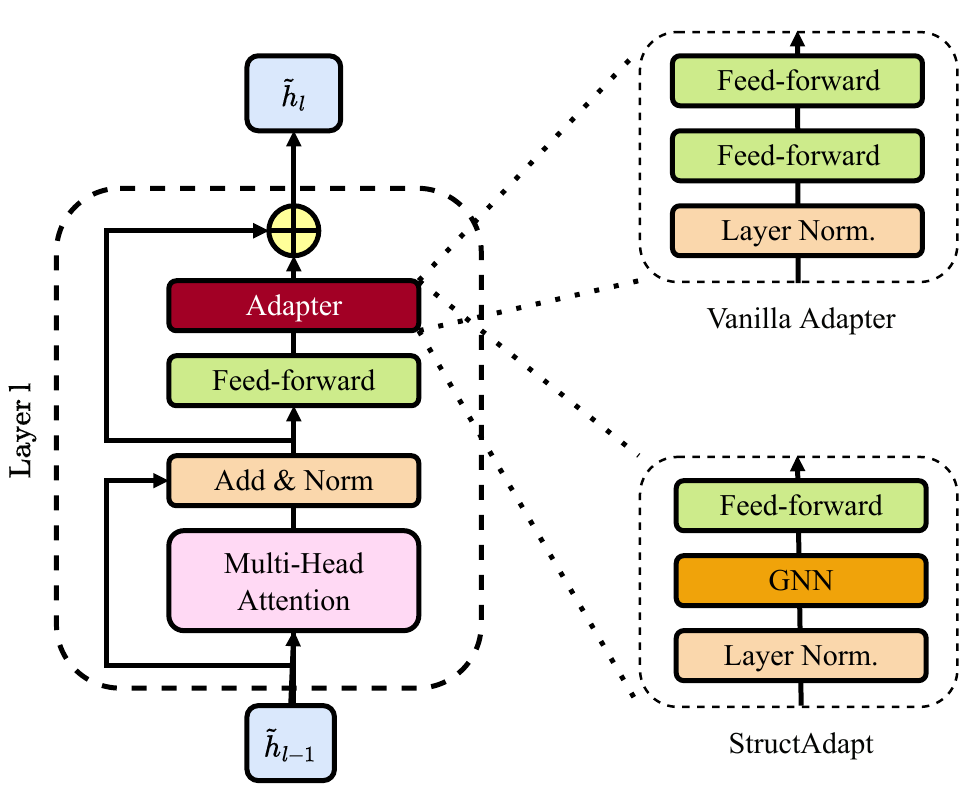}
    \caption{Vanilla Adapter vs StructAdapt}
    \label{fig:adapters}
\end{figure}

A major issue in AMR-to-Text, and more generally Graph-to-Text with Transformers \cite{NIPS2017_3f5ee243}, is the linearization of the input structure. The linearization of the graph returns a sequence of node and edge labels according to a certain traversal of the graph. Nonetheless, adjacent nodes in the graph may be at multiple positions away from one another in the final serialization. 
To counteract this, \citet{ribeiro-etal-2021-structural} introduced StructAdapt, a structure-aware (encoder) adapter. It solves the problem of segmented nodes labels by reconstructing a new graph from the resulting subwords. More specifically, the relations are primary reified as new nodes in the AMR graph. 
Furthermore, labels of those reified relations will be added in the vocabulary as new tokens and therefore will not be decomposed into subwords. However, the labels of the original nodes can still be chunked. To deal with this, each subword node will be connected independently to the reified relation of the initial (non-chunked) node. 
An example is outlined in Figure \ref{fig:tokenized_amr}.
As a consequence, the vanilla Adapter can now integrate any GNN-based neural network which operates on the new constructed graph (Figure \ref{fig:tokenized_amr}), where nodes are the input tokens. Concretely, StructAdapt replaces the first stacked MLP of vanilla adapter with a GNN-based model as shown in Figure \ref{fig:adapters}. 
\begin{table*}[h!]
\centering
\scriptsize

\begin{tabular}{ccccccccc}
\multicolumn{2}{c}{\textbf{Adapter}} & \textbf{BLEU $\uparrow$} & \textbf{METEOR $\uparrow$} & \textbf{Chrf++ $\uparrow$} & \textbf{TER $\downarrow$} &  \textbf{BERTScore $\uparrow$} & \textbf{$\mathcal{M}\uparrow$} & \textbf{$\mathcal{F}\uparrow$} \\ \hline
\multirow{2}{*}{\textbf{MLP}} & \textbf{w/ \, RPE} & \textbf{41.6}\tiny{$\pm$0.3} & \textbf{56.5}\tiny{$\pm$0.3} & \textbf{70.6}\tiny{$\pm$0.2}  & \textbf{45.9}\tiny{$\pm$0.5}  & \textbf{95.6}\tiny{$\pm$0.0}  & 
\textbf{84.2}\tiny{$\pm$0.1} & 
\textbf{78.0}\tiny{$\pm$0.5} \\
 & \textbf{w/o RPE} & 16.3\tiny{$\pm$0.5} & 38.5\tiny{$\pm$0.8}  & 49.9\tiny{$\pm$1.1}  & 85.5\tiny{$\pm$0.9}  & 91.8\tiny{$\pm$0.2} & 81.9\tiny{$\pm$1.1} & 76.2\tiny{$\pm$0.6} \\ \hline
\multirow{2}{*}{\textbf{GCN}} & \textbf{w/ \, RPE} & \textbf{42.6}\tiny{$\pm$0.8} & \textbf{56.7}\tiny{$\pm$0.4} & \textbf{71.0}\tiny{$\pm$0.5} & \textbf{44.8}\tiny{$\pm$0.7}  & \textbf{95.7}\tiny{$\pm$0.1} & 
\textbf{84.6}\tiny{$\pm$0.3} & 
\textbf{79.0}\tiny{$\pm$0.6} \\
 & \textbf{w/o RPE} & 
 34.4\tiny{$\pm$0.8} & 52.0\tiny{$\pm$0.7} & 64.8\tiny{$\pm$0.6}  & 55.8\tiny{$\pm$1.2}  & 94.6\tiny{$\pm$0.1} &
 79.2\tiny{$\pm$0.6} & 
 75.2\tiny{$\pm$1.0}  \\ \hline
\multirow{2}{*}{\textbf{GAT}} & \textbf{w/ \, RPE} & \textbf{42.8}\tiny{$\pm$0.1} & \textbf{57.0}\tiny{$\pm$0.1} & \textbf{71.1}\tiny{$\pm$0.0} & \textbf{44.3}\tiny{$\pm$0.3}  & \textbf{95.8}\tiny{$\pm$0.0}  & 
\textbf{84.8}\tiny{$\pm$0.1} & 
\textbf{78.5}\tiny{$\pm$0.8} \\
 & \textbf{w/o RPE} & 34.8\tiny{$\pm$1.1} & 52.3\tiny{$\pm$0.7} & 64.8\tiny{$\pm$0.7} & 54.9\tiny{$\pm$1.4}  & 94.6\tiny{$\pm$0.2}  & 79.6\tiny{$\pm$0.8} & 
 75.6\tiny{$\pm$0.4} \\ \hline
\multirow{2}{*}{\textbf{RGCN}} & \textbf{w/ \, RPE} & \textbf{44.7}\tiny{$\pm$0.6} & \textbf{58.2}\tiny{$\pm$0.3} & \textbf{72.5}\tiny{$\pm$0.3} & \textbf{42.6}\tiny{$\pm$0.4}  & \textbf{96.0}\tiny{$\pm$0.0}  & 
\textbf{85.5}\tiny{$\pm$0.2} & 
\textbf{79.6}\tiny{$\pm$0.7} \\
 & \textbf{w/o RPE} & 39.9\tiny{$\pm$0.8} & 55.7\tiny{$\pm$0.5} & 68.9\tiny{$\pm$0.8}  & 48.8\tiny{$\pm$1.3}  & 95.3\tiny{$\pm$0.1} & 
 83.1\tiny{$\pm$0.4} & 
 78.0\tiny{$\pm$0.6} \\ \hline
\end{tabular}
\caption{\label{table:with_without_rpe_correct_coo}
Comparing impact of Relative Positional Embeddings (RPE) on generation. We report mean performances ($\pm$s.d.) over 3 seeds.
}
\end{table*}
\begin{table}
\centering
\scriptsize
\renewcommand{\arraystretch}{1.3}
\begin{tabular}{ccc}
\hline
\textbf{Adapter} & \begin{tabular}[c]{@{}c@{}}\textbf{Meaning} \\ \textbf{Preservation}\end{tabular} & \begin{tabular}[c]{@{}c@{}}\textbf{Linguistic} \\ \textbf{Correctness}\end{tabular} \\ \hline
\textbf{MLP w/  RPE} & \textbf{4.8}\tiny{$\pm$1.2} & \textbf{5.5}\tiny{$\pm$0.9} \\ 
\textbf{MLP w/o  RPE} & 3.5\tiny{$\pm$1.5} & 5.2\tiny{$\pm$1.2} \\ \hline
\textbf{GCN w/ RPE} & \textbf{5.0}\tiny{$\pm$1.2} & \textbf{5.6}\tiny{$\pm$0.8} \\
\textbf{GCN w/o RPE} & 4.7\tiny{$\pm$1.3} & 5.5\tiny{$\pm$1.0} \\  \hline
\textbf{RGCN w/ RPE} & \textbf{5.2}\tiny{$\pm$1.1} & \textbf{5.6}\tiny{$\pm$0.8} \\
\textbf{RGCN w/o RPE} & 4.7\tiny{$\pm$1.3} & 5.4\tiny{$\pm$1.0} \\ \hline
\end{tabular}
\caption{Human Evaluation. Mean scores ($\pm$s.d.)}
\label{table:human_evaluation}
\end{table}
For AMR-to-Text, only the encoder is equipped with StructAdapt in order to encode AMR structure. The decoder layers adopt vanilla adapters. 
In our study, we consider three different GNN-based models which are Graph Convolutional Network (GCN) \cite{kipf-etal-2016-gcn},  Graph Attention network (GAT) \cite{velikovi-etal-2017-gat} and Relational Graph Convolutional Network (RGCN) \cite{Schlichtkrull-etal-2018}. GCN computes a representation for each node $a$ which is a (normalized) aggregation function of representation of its neighbor nodes noted $\mathcal{N}(a)$. GAT is akin to GCN but differs in that aggregation of neighbors embeddings are weighted using an attention mechanism. Unlike GAT and GCN, RGCN further captures the type of the relation between two nodes. In our case, as AMR relations are reified and stand for new nodes of the graphs, our new relations can either be of \textit{direct} ($a\xrightarrow[]{d}b$) or \textit{reverse} ($a\xleftarrow[]{r}b$) connections type as in \cite{ribeiro-etal-2021-structural}. 
The details of the representations computation for each model can be found in Appendix \ref{sec:appendix_gnn}. The returned nodes embeddings will then be given as input features to the following MLP.
\section{Relative Position Embeddings}
Instead of adding Absolute Position Embeddings (APE) directly to the token embedding as in standard Transformer model, some models such as T5 make use of relative position embeddings inspired from \citet{shaw-etal-2018-self}. 
As an alternative to APE, RPE offer interesting features. A noteworthy limitation of APE is the need to set a limit of available positions. Therefore, long sequences may have to be segmented. Furthermore, APE are directly added to the token representation leading to information inconsistency, i.e. position versus semantic information. To this extent, \citet{shaw-etal-2018-self} came up with relative position encodings which are supplied to the self-attention mechanism by simply adding a scalar to the logits encoding the supposed relation between a current token $i$ and a token $j$. 
\section{Experiments}
Throughout our experiments, we make use of the LDC2020T02 dataset (AMR 3.0 release)\footnote{https://catalog.ldc.upenn.edu/LDC2020T02} and use the T5$_{{\rm base}}$ model which employs RPE. The training and evaluation details can be found in Appendix \ref{sec:appendix_training_settings} and \ref{sec:appendix_automatic_evaluation}, respectively. 
\subsection{Exploring the Salience of RPE}
\label{section:structadapt_robustness}
\begin{table*}[h!]
\centering
\scriptsize
\begin{tabular}{ccccccccc}
\multicolumn{2}{c}{\textbf{Adapter}} & \textbf{BLEU $\uparrow$} & \textbf{METEOR $\uparrow$} & \textbf{Chrf++ $\uparrow$} & \textbf{TER $\downarrow$} &  \textbf{BERTScore $\uparrow$} & \textbf{$\mathcal{M}\uparrow$} & \textbf{$\mathcal{F}\uparrow$} \\ \hline
\multirow{2}{*}{\textbf{GCN}} & \textbf{w/ \, RPE} & \textbf{35.1}\tiny{$\pm$0.5} & \textbf{52.0}\tiny{$\pm$0.4} & \textbf{65.7}\tiny{$\pm$0.5} & \textbf{53.6}\tiny{$\pm$0.9}  & \textbf{94.8}\tiny{$\pm$0.1} &  
\textbf{80.5}\tiny{$\pm$0.4} & 
74.1\tiny{$\pm$1.2} \\
 & \textbf{w/o RPE} & 13.9\tiny{$\pm$0.5} & 35.2\tiny{$\pm$0.6} & 46.7\tiny{$\pm$0.5}  & 86.5\tiny{$\pm$2.7}  & 91.1\tiny{$\pm$0.1} &  77.1\tiny{$\pm$14.4} & 
 \textbf{78.0}\tiny{$\pm$2.6} \\ \hline
\multirow{2}{*}{\textbf{GAT}} & \textbf{w/ \, RPE} & \textbf{36.8}\tiny{$\pm$0.6} & \textbf{52.3}\tiny{$\pm$0.5} & \textbf{67.0}\tiny{$\pm$0.5} & \textbf{51.1}\tiny{$\pm$0.9}  & \textbf{95.0}\tiny{$\pm$0.1}  &  \textbf{81.6}\tiny{$\pm$0.5} & 
 75.7\tiny{$\pm$0.3} \\
 & \textbf{w/o RPE} & 13.1\tiny{$\pm$2.8} & 34.1\tiny{$\pm$3.6} & 45.2\tiny{$\pm$4.2} & 89.1\tiny{$\pm$0.6}  & 90.7\tiny{$\pm$0.1}  &  68.9\tiny{$\pm$14.1} & 
 \textbf{77.4}\tiny{$\pm$1.9} \\ \hline
\multirow{2}{*}{\textbf{RGCN}} & \textbf{w/ \, RPE} & \textbf{38.0}\tiny{$\pm$0.9} & \textbf{54.1}\tiny{$\pm$0.6} & \textbf{67.6}\tiny{$\pm$0.6} & \textbf{49.3}\tiny{$\pm$0.8} & \textbf{95.2}\tiny{$\pm$0.1} & 
\textbf{81.9}\tiny{$\pm$0.7} & 
 \textbf{75.9}\tiny{$\pm$0.8} \\
 & \textbf{w/o RPE} & 11.3\tiny{$\pm$1.3} & 30.1\tiny{$\pm$1.1} & 41.6\tiny{$\pm$1.0}  & 87.2\tiny{$\pm$1.3}  & 90.0\tiny{$\pm$0.2} &  66.8\tiny{$\pm$16.3} & 
 75.5\tiny{$\pm$5.2} \\ \hline
\end{tabular}
\caption{\label{table:graph_attack}
\textit{Graph Attack} - We corrupt the graph connectivity information given to the structural adapters in encoder. We report mean performances ($\pm$s.d.) over 3 seeds. 
}
\end{table*}
\begin{table}
\centering
\scriptsize
\renewcommand{\arraystretch}{1.3}
\begin{tabular}{ccc}
\hline
\textbf{Adapter} & \begin{tabular}[c]{@{}c@{}}\textbf{Meaning} \\ \textbf{Preservation}\end{tabular} & \begin{tabular}[c]{@{}c@{}}\textbf{Linguistic} \\ \textbf{Correctness}\end{tabular} \\ \hline
\textbf{GCN w/ RPE} & 5.0\tiny{$\pm$1.2} & 5.6\tiny{$\pm$0.8} \\
\textbf{RGCN w/ RPE} & 5.2\tiny{$\pm$1.1} & 5.6\tiny{$\pm$0.8} \\ \hline
\end{tabular}
\caption{\textit{Graph Attack} - Human Evaluation. Mean scores ($\pm$s.d.)}
\label{table:graph_attack_human_evaluation}
\end{table}
In this section, we investigate the influence of RPE on the generation quality using structural adapter. RPE are computed in each Transformer head. Position information is then forwarded to the adapter module on top (Figure \ref{fig:adapters}). However, since connections between input nodes (i.e. tokens) are already given to structural adapters in encoder, it is legitimate to question the necessity for RPE on the encoder part but also how would the generation quality vary without such information. Hence, we propose to remove the RPE from the encoder heads to gauge their salience to structure encoding for downstream language generation. Since decoder is not encoding any graph structure, we leave RPE in decoder untouched. For better readability, MLP, GCN, GAT, and RGCN respectively denote: the vanilla adapter, StructAdapt with a GCN layer, StructAdapt with a GAT layer and StructAdapt with a RGCN layer. MLP-based adapter with RPE is our baseline. Results are given in Table \ref{table:with_without_rpe_correct_coo}. A human evaluation is also provided for some encoder adapters in Table \ref{table:human_evaluation}. 
First, it is apparent that using RPE systematically yields better generation performances. For the vanilla adapter (i.e.\ our baseline), we note a 25.3\% absolute drop in BLEU when removing RPE. This can also be seen on human evaluation. More than one point is lost toward meaning preservation. The downturn for linguistic correctness is less important since T5 is pretrained and thus rarely prone to syntax errors. Such a result is not surprising for MLP-based adapter since it solely relies on RPE to differentiate tokens at different positions in the linearized AMR.  However, a striking observation is that getting rid of RPE for GNN-based adapters leads to lower performances than our baseline. Indeed, when removing RPE when using structural adapter, we would have expected GNN-based approaches to be as competitive as a MLP-based adapter with RPE. We report a relative drop of 12.5 points in BLEU from the baseline. The same conclusion can be drawn from Table \ref{table:human_evaluation}. This indicates that RPE are capturing relevant information for final generation.
To further assess the impact and the role of RPE, we conduct a \textit{graph attack} experiment. Instead of conveying the correct adjacency matrix, we propose to corrupt connectivity information. We randomly generate an adjacency matrix such that generated matrix does not contain any actual connection. 
We suppose that without RPE, structure-aware adapter will lead to a significant decrease in generated text due to the absence of information about the graph nor the position of nodes in the input sequence. We are especially interested to measure to which extent RPE might be able to take over the encoding of the graph for generation. Results are shown in Table \ref{table:graph_attack}. Human evaluation for \textit{attacked} GCN and RGCN adapters is given in Table \ref{table:graph_attack_human_evaluation}. As hypothesized, providing erroneous connectivity without any position embeddings makes structural adapters no longer compelling. We observe that StructAdapt with RGCN is significantly more affected compared to GAT and GCN based adapters. Since RGCN adds direction information for each edge (\textit{direct} and \textit{reverse}), we conjecture that RGCN is much more bewildered. Interestingly, using RPE with corrupted graph (Table \ref{table:graph_attack}) leads to similar performance than using graph information without RPE (Table \ref{table:with_without_rpe_correct_coo}). This strongly demonstrates the usefulness of RPE to carry out the generation. We additionally provide a \textit{position attack} experiment in Appendix \ref{sec:appendix_position_attack} where RPE are shuffled randomly. Accordingly, we can further identify the saliency of RPE despite the available GNN. This raises the question of RPE encoding the input graph.

\subsection{Can the Graphs Be Reconstructed ?}
\begin{figure}
    \centering
    \includegraphics[width=\linewidth]{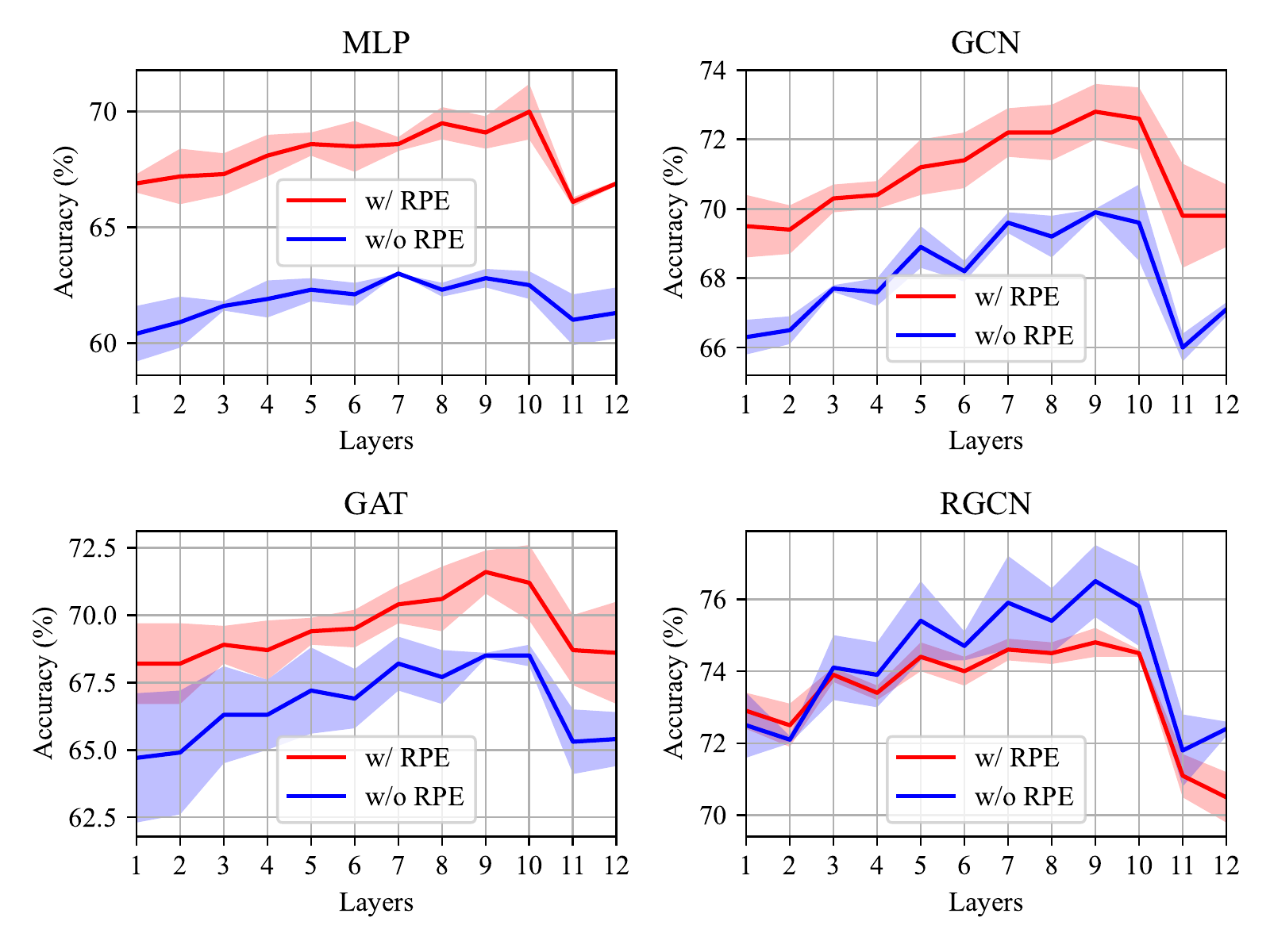}
    \caption{\textit{Probing} - Link Prediction Results using hidden representations $\Tilde{h}_{i}$  at different layers $i$.}
    \label{fig:link_prediction_plot}
\end{figure}

As shown in Section \ref{section:structadapt_robustness}, RPE seem to be as competitive as applying GNNs alone. If claiming that RPE also encode graphs is tempting, no strong evidence has been revealed. Indeed, better generation quality is not necessarily a consequence of better graph encoding. Therefore, we probe whether graphs can indeed be reconstructed from the learned hidden representations. To do so, we train a logistic regression, i.e. our probe, to perform link prediction as a binary classification. More specifically, given two nodes representations at a given layer $l$, our probe returns the probability that nodes are connected. To train our logistic regression model, we sample $k$ positive connections, i.e. two connected nodes,  and $k$ negative connections, i.e. two non-connected nodes, for each sample of training and test sets.\footnote{If $k$ samples cannot be extracted for both positive and negative classes, the sample is discarded.} For our experiment, we choose $k=2$ which leads to 109,490 and 3,770 samples for each class for training and testing respectively. We plot results in Figure \ref{fig:link_prediction_plot}. 
Firstly, we observe very high accuracy for the vanilla adapter without RPE. Accuracies over 60\% are easily reached while no structure encoding nor positions information are supplied. This might be a side effect of our probe training. Nevertheless, this gives a lower bound for our experiment. We can see that adding RPE increases the link prediction performance for MLP, GCN and GAT-based adapters. We observe a constant gap of about 3.5\% on average. However, we remark that RGCN is able to reconstruct edges on its own with best accuracy. We report a maximum accuracy about 76\% while other models are not reaching 73\%. This strengthens the idea that giving information of reverse connection may add robustness to graph encoding as shown in \cite{beck-etal-2018-graph}. Generally, we observe that the deeper the representation, the better the link prediction. We notice however that after the 10$^{th}$ layer, significant drops in link prediction arise regardless of adapter type. We assume representations should lose some information about the structure to perform language generation. This may indicate that encoded representation for Graph-to-Text is not just graph-centered. Although counter-intuitive, encoder representations given to the decoder part may not have to encode input graph efficiently in order to verbalize it. We leave this research question for future work. We further provide an analysis on self-attention matrices in Appendix \ref{sec:appendix_self_attention}.

\section{Conclusion}
In this paper, we have explored the effect of relative position embeddings on AMR-to-Text generation using structural adapters.  We have shown that the generation process could be enabled by relative position embeddings when structure is erroneous or missing. In addition, we have demonstrated the capacity of those representations to encode the input graph to some extent. We have further revealed interesting robustness of RGCN model in graph reconstruction ability. For future work, we believe further experiments on other pretrained models and Graph-to-Text tasks may shed more light on the role of position embeddings.

\section*{Limitations}
A limitation of our study is that we focus on the T5 model only. Since adapters are additional modules to add, it is required to manually implement and directly modify the original code of the pretrained model which is not easily scalable. In addition, we only evaluate on the LDC2020T02 dataset which is the cleanest AMR dataset available.   

\section*{Acknowledgements}
We would like to thank annotators and reviewers for taking the time and effort necessary to share our contribution. This work was partially funded by the ANR Cifre conventions N°2020/0400 and  Orange Innovation Research.

\bibliography{anthology,custom}
\bibliographystyle{acl_natbib}

\appendix

\section{Graph Neural Networks}
\label{sec:appendix_gnn}
\subsection{Graph Convolutional Network (GCN)}
To compute the node representation $g_{u}^{l} \in \mathbb{R}^{d_{g} }$ of the node $u$ at layer $l$, GCN computes an aggregation of neighbors nodes embeddings as: 
\begin{equation}
    g_{u}^{l} = \sigma \left(\sum_{v \in \mathcal{N}(u)} \frac{1}{\text{deg}(u)\times \text{deg}(v)} W^{l}g_v^{l} \right)
\end{equation}
with $\sigma$ an activation function and ${\rm deg}(x)$ the degree of the node $x$.
\subsection{Graph Attention Network (GAT)}
GAT applies an attention mechanism to determine importance of neighboors nodes regarding current node $u$. We have: 
\begin{equation}
\begin{split}
    g_{u}^{l} = \sigma \left( \sum_{v \in \mathcal{N}_(u)} {\rm softmax}({e_{u, v})} W^{l}g_v^{l} \right) \\
    e_{u, v} = {\rm LeakyReLu}\left( a_{u, v} \underset{i=u}{\overset{v}{\bigg\Vert g_i^{l}}}\right)
\end{split}
\end{equation}
with both $\sigma$ and ${\rm LeakyReLu}$ non-linear activation functions and $a_{u, v} \in \mathbb{R}^{2\times d_{g}}$ a learnable vector.

\subsection{Relational Graph Convolutional Network (RGCN)}
RGCN takes into consideration the nature of the relation $r \in \mathcal{R}$ between nodes $u$ and $v$. It performs convolution as the following:
\begin{equation}
    g_{u}^{l} = \sigma \left(\sum_{r\in \mathcal{R}}\sum_{v \in \mathcal{N}_{r}(u)} c_{u, r} W^{l}_{r}g_v^{l} \right)
\end{equation}
with $\mathcal{N}_{r}(u)$ the direct neighbors of node $u$ under the relation $r$, $c_{u, r}$ a normalization term and $W^{l}_{r} \in \mathbb{R}^{d_{g} \times d_{g}}$ a learnable relation-dependent parameter.

\section{Training Setting}
\label{sec:appendix_training_settings}
We used the version 3.3.1 of the HuggingFace’s Transformers library \cite{wolf-etal-2020-transformers}. We use Adam optimizer with a linearly decreasing learning rate, without warm-up. The initial learning rate is set to $1\times 10^{-4}$. Batch size is set to 8. For decoding, a beam search of 5 is selected. A maximum length of 384 is used in case the \textit{end-of-sentence} token is not encountered. We didn't use gradient clipping nor label smoothing. For GNNs, we make use of the version 1.7.2 of the PyTorch Geometric library \cite{Fey/Lenssen/2019}. We only use a single layer for our GNNs networks, similarly to the vanilla adatper. For GAT, a single attention head was applied. The bottleneck dimension is set to 256 for all of our adapters. This is equivalent to about 4\% only of the whole T5$_{base}$ model's parameters to train. The training time for MLP, GCN, GAT and RGCN adapters are given in Table \ref{table:runtime}. Note that the total runtime depends on the convergence as we are using early stopping.
\begin{table}
\centering
\scriptsize
\renewcommand{\arraystretch}{1.3}
\begin{tabular}{ccc}
\hline
\textbf{Adapter} & \begin{tabular}[c]{@{}c@{}}\textbf{Runtime} \\ \textbf{(in hours)}\end{tabular} \\ \hline
\textbf{MLP} & 21.5 \\
\textbf{GCN} & 15.6 \\
\textbf{GAT} & 22.0 \\
\textbf{RGCN} & 16.2 \\ \hline
\end{tabular}
\caption{\textit{Runtime} - Runs are executed on a single NVIDIA GeForce RTX 3090 GPU.}
\label{table:runtime}
\end{table}

\section{Automatic Evaluation}
\label{sec:appendix_automatic_evaluation}
For evaluation, we consider multiple automated metrics. We employ popular $n$-gram-based metrics which are SacreBLEU \cite{post-2018-call}, METEOR \cite{banerjee-lavie-2005-meteor}, chrf++ \cite{popovic-2017-chrf}, TER \cite{och-2003-minimum} and the semantic-based BertScore metric \cite{bertscore}. In addition, we also report both $\mathcal{M}$ and $\mathcal{F}$ pillars of the decomposable $\mathcal{M}\mathcal{F}$ score recently proposed by \citet{opitz-frank-2021-towards}. The meaning preservation, noted $\mathcal{M}$, measures how close the AMR of the generated sentence is to the reference sentence. To do so, both the generated sentence and target sentence are parsed with a pretrained Text-to-AMR model from \citet{cai-lam-2020-amr}. Then, their AMRs are compared using graph-based similarity measures such as Smatch \cite{cai-knight-2013-smatch}.  
In contrast, the grammatical form, noted $\mathcal{F}$, measures the linguistic quality of the generated text using state-of-the-art language model \cite{Radford2019}. 
We also report human evaluation in Section \ref{section:structadapt_robustness} to accurately assess the quality of the generated outputs. We ask annotators to evaluate the meaning preservation and linguistic correctness of our generated outputs compared to the references. Details of human evaluation are given in Appendix \ref{sec:appendix_human_evaluation}.

\section{Human Evaluation}
\label{sec:appendix_human_evaluation}
Since handcrafted annotation is extremely costly, we limited the number of samples and models to assess. We randomly selected 30 test samples per model to evaluate manually.  We further limited evaluation to 8 configurations of adapters. This led us with 240 samples to score by multiple annotators. We asked 22 annotators to answer to these following questions with a rate on a 1-6 Likert scale, with 6 the best score:
\begin{itemize}
    \item Is the generated sentence semantically close to the reference ? 
    \item Is the generated sentence grammatically correct ? 
\end{itemize}
Each annotator was given 50 samples to annotate. We adopted FlexEval \cite{fayet-etal-2020-flexeval} to implement a flexible evaluation environment. This allows a streaming evaluation were annotators can stop at any moment and come back to the evaluation later, at one's will. Furthermore, different samples from different models are given to annotators in an balanced manner such that each sample of each model is exposed to at least 2 annotators. Note that one annotator is given some samples from multiple configuration. This avoids evaluation bias toward a single annotator. In our case, each sample from each configuration has been annotated by 4 distinct annotators.

\section{Position Attack}
\label{sec:appendix_position_attack}
For a given graph, we also propose to corrupt RPE with a random shuffle. Since the model may learn the generated order through training, we \textit{systematically} shuffle RPE for each epoch for all graphs. Results are given in Table \ref{table:position_attack}.
\begin{table}[h!]
\centering
\scriptsize
\renewcommand{\arraystretch}{1.3}
\begin{tabular}{ccc}
\hline
\textbf{Adapter} & \textbf{BLEU} \\ \hline
\textbf{MLP} & 19.0\tiny{$\pm$0.6} \\
\textbf{GCN} & 28.1\tiny{$\pm$0.6} \\
\textbf{GAT} & 28.9\tiny{$\pm$2.3} \\
\textbf{RGCN} & \textbf{35.4}\tiny{$\pm$1.2}\\ \hline
\end{tabular}
\caption{\textit{Position Attack} - Relative Position Embeddings are shuffled.}
\label{table:position_attack}
\end{table}

\begin{figure*}
    \centering
    \includegraphics[width=0.8\linewidth]{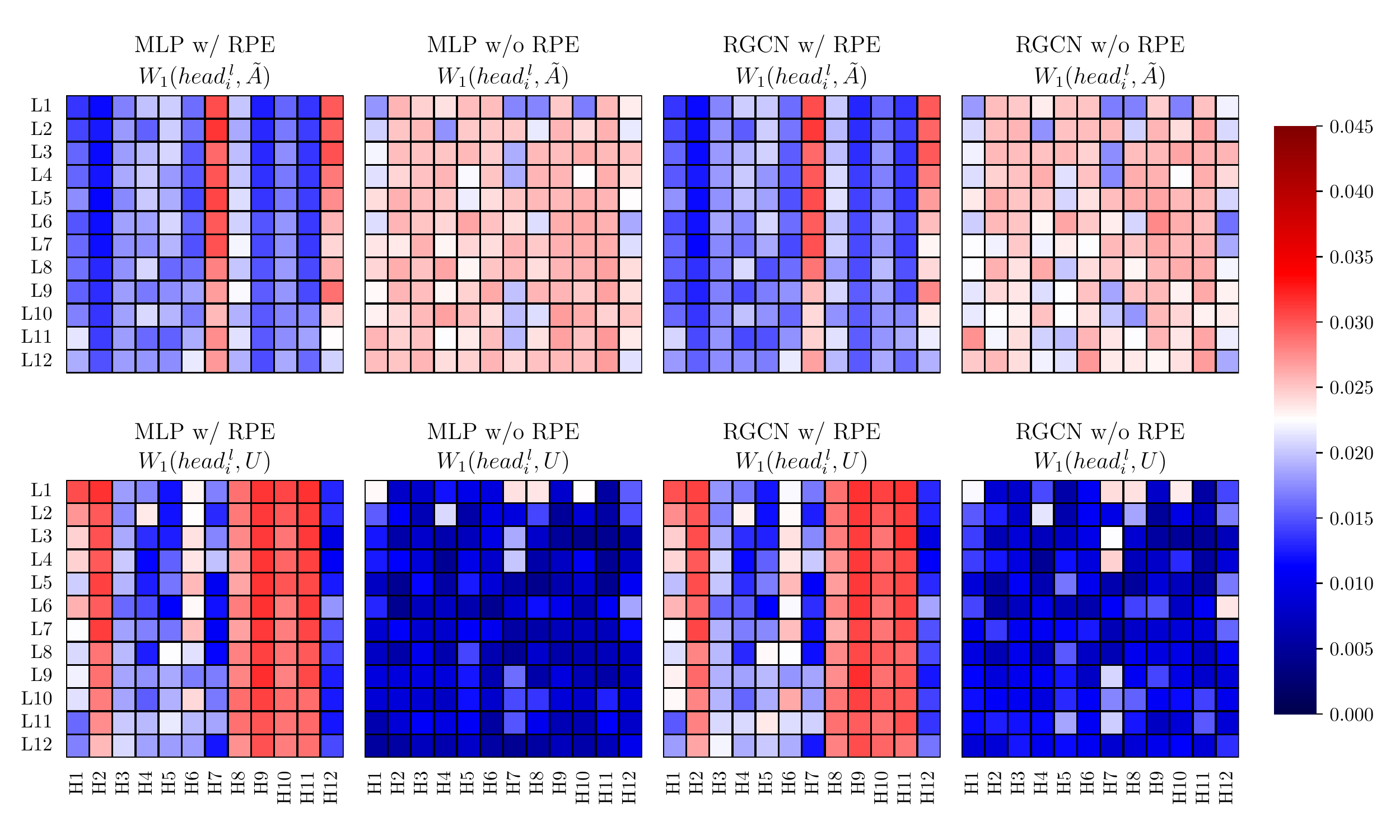}
    \caption{1-Wasserstein distances of attention distributions for each head (columns) at each layer (rows) from both normalized adjacency matrices (top) and uniform distribution (down). The bluer, the lower the distance.}
    \label{fig:wasserstein}
\end{figure*}

\section{Analysis of Attention Matrices}
\label{sec:appendix_self_attention}
The self-attention matrix gives information about how much weight should be assigned to other tokens (i.e. nodes). Thus, attention matrix can be seen as a pseudo-adjacency matrix. We further propose to inspect the similarity between the attention matrix of each head at each layer with the adjacency matrix of the input graph. Since the input is considered as a sequence of nodes, the shape of the attention matrix and adjacency matrix are equal. However, a limitation of the similarity comparison lies in the respective type of those mathematical objects. Attention matrices are probabilities whereas adjacency matrices are not. To deal with this issue, we convert the adjacency matrix $A^{s}$ of a sample $s$ to a normalized form $\Tilde{A}^{s}$ such that each row $\Tilde{A}^{s}_{n}$ is normalized and sums up to one, as described in Eq. \ref{eq:normalize_adj}. 
\begin{equation}
A^{s}=\begin{bmatrix}
0 & 1 & 1 \\[1pt]
1 & 1 & 1 \\[1pt]
0 & 0 & 1
\end{bmatrix} \quad
\Tilde{A}^{s}=\begin{bmatrix}
0 & \frac{1}{2} & \frac{1}{2} \\[1pt]
\frac{1}{3} & \frac{1}{3} & \frac{1}{3} \\[1pt]
0 & 0 & 1
\end{bmatrix}
\label{eq:normalize_adj}
\end{equation}
This enables the rows of both attention and adjacency matrices to be probabilities distributions over input nodes. We can therefore measure the distance between those distributions to find out whether attention matrices are somehow close to the (normalized) adjacency matrix, and thus encoding graph connections. To do so, we compute the 1-Wasserstein distance $W_{1}$ \footnote{Equivalent to earth mover's distance.} between the attention distribution of each token $n$ and its corresponding normalized $n^{th}$ row of $\Tilde{A}$. We average distances over tokens, and then over samples (Eq. \ref{eq:wassertein}). We note $\Tilde{A}$, $S$, and $N_{s}$ the adjacency matrices, the total number of AMR samples and the length of sequence $s$, respectively.
\begin{equation}
\begin{split}
    W_{1}({\rm head}^{l}_{i}, \Tilde{A}) =\\ \frac{1}{S}\sum_{s=1}^{S}&\frac{1}{N_{s}}\sum_{n=1}^{N_{s}}W_{1}({\rm head}_{i, n}^{l, s}, \Tilde{A}_{n}^{s})
\end{split}
\label{eq:wassertein}
\end{equation}
For fair comparison, we also provide the distance between attention distribution ${\rm head}^{l}_{i}$ and the uniform distribution ${U}$ which assigns a probability of $\frac{1}{N}$ to each token for a graph with $N$ nodes. We plot distances as heatmaps on Figure \ref{fig:wasserstein} where each square indicates either $W_{1}({\rm head}^{l}_{i}, \Tilde{A})$ or $W_1({\rm head}^{l}_{i}, {U})$.
When using RPE, we can see that, for both MLP and RGCN-based adapters, the distribution of the attention scores are close to the normalized adjacency matrices of our graphs. Meanwhile, when removing them, we can witness that attention scores are much closer to the uniform distribution and thus smoother. This is in line with our previous results suggesting that RPE might partially encode graphs. Since RPE are shared across layers, we can easily detect similar distances across layers for same heads (visible vertical lines). We also find similar behavior we have previously noted where last layers tend to lose graph information as attention scores slightly move away from the normalized adjacency matrix distribution. 

\end{document}